# Large Scale Multimodal Classification Using an Ensemble of Transformer Models and Co-Attention

Varnith Chordia
vchordia@parc.com
Palo Alto Research Center
Palo Alto, California

Vijay Kumar
vkumar@parc.com
Palo Alto Research Center
Palo Alto, California

## ABSTRACT
Accurate and efficient product classification is significant for E-commerce applications, as it enables various downstream tasks such as recommendation, retrieval, and pricing. Items often contain textual and visual information, and utilizing both modalities usually outperforms classification utilizing either mode alone. In this paper we describe our methodology and results for the SIGIR eCom Rakuten Data Challenge. We employ a dual attention technique to model image-text relationships using pretrained language and image embeddings. While dual attention has been widely used for Visual Question Answering(VQA) tasks, ours is the first attempt to apply the concept for multimodal classification.

## 1 INTRODUCTION
The E-commerce business has seen tremendous growth over the last decade, with online shopping offering an ever-expanding range of products. Given the explosive growth of inventories, there is a growing need for automated product classification that can enable and streamline downstream tasks such as search, retrieval, and recommendations

The SIGIR eCom Rakuten Data Challenge poses two tasks: 1.product classification and 2.cross modal retrieval. In this paper we focus on the first task, namely the classification of large-scale multimodal (text and image) product data into product type codes in the catalog of Rakuten France.

In this paper we employ multilingual and monolingual transformer language models to generate rich text representation, eliminating the need for higher order text features. We combine language and visual representations using a modified version of the co-attention architecture proposed in Lu et al. [11]. We adopt a stacking based ensemble approach that combines models from different experiments to improve F1-score. Experiments show that our technique outperforms a baseline approach of straightforward classification that fuses text and image representations. We have made our code publicly available for use[1].

## 2 RELATED WORK
Visual Question Answering (VQA) [3] is a multimodal learning task that lies at the intersection of image and text processing. The most straightforward architecture [3] uses a combination of VGGNet[17] and LSTM[8]. Several methods [2, 14] construct an architecture focused on the question. A drawback of these methods is that they consider only global image context, which may contain information irrelevant to the question. To overcome this, some methods have proposed visual attention models that attend to local spatial regions pertaining to a given question, and then perform multimodal fusion to classify answers accurately [4, 19, 21, 22]. More recently, dual attention models have been proposed. The work by Nam et al. [13] jointly reasons about visual and textual attention mechanisms to capture the fine grained interaction between image and text, thereby achieving higher performance. We build on the co-attention approach developed by Lu et al. [11]. Most approaches have focused on using glove embeddings [15] to encode text based features, but not much work has been done on using transformer language models.

Bidirectional Encoder Representations from Transformers (BERT) is an open-source state of art pre-trained language model [5], [1]. BERT models are frequently fine tuned for specific tasks. Sun et al. [18] demonstrated improved performance in text classification by using different strategies such as layer selection, layer wise learning rate, etc. Many variations of BERT have produced advances in language processing tasks. FlauBERT [10] and CamemBERT [12] are example of variations of BERT trained on large French corpora.

## 3 METHODOLOGY
In this section, we describe our baseline approach and proposed model architecture for product type classification. Our general approach is to project the unimodal representations from image ($X^i$) and text ($X^t$) into a joint representation space and learn a classifier in this joint space. i.e,

$$y = f(Z(X^i, X^t)) \quad (1)$$

where $Z$ is function (learned or hand-crafted) that combines the text and image features, and $f$ is a neural network classifier that maps the features from the joint feature space to a class label $y$.

### 3.1 Baseline Approach
For the baseline, we use the deep features extracted from pretrained image and text models, and explore several techniques to fuse the features from these two modalities. We then learn a small network with two fully connected layers followed by a softmax classifier to predict the product type.

In detail, we encode text for every product through CamemBERT which outputs embeddings for every token in the text across a 12 layered hidden state. The output is represented as $W^{12 \times n \times d}$ where n is the number of tokens and d is the embedding dimension. The embeddings of the first token of the top layer are selected to produce a single dimension embedding representation of the text, $W^d$. For image features, we use 2048-d top layer pooling activations from ResNet152 that is pretrained on ImageNet-1K.

---
[1]https://github.com/VarnithChordia/Multimodal_Classification_Co_Attention

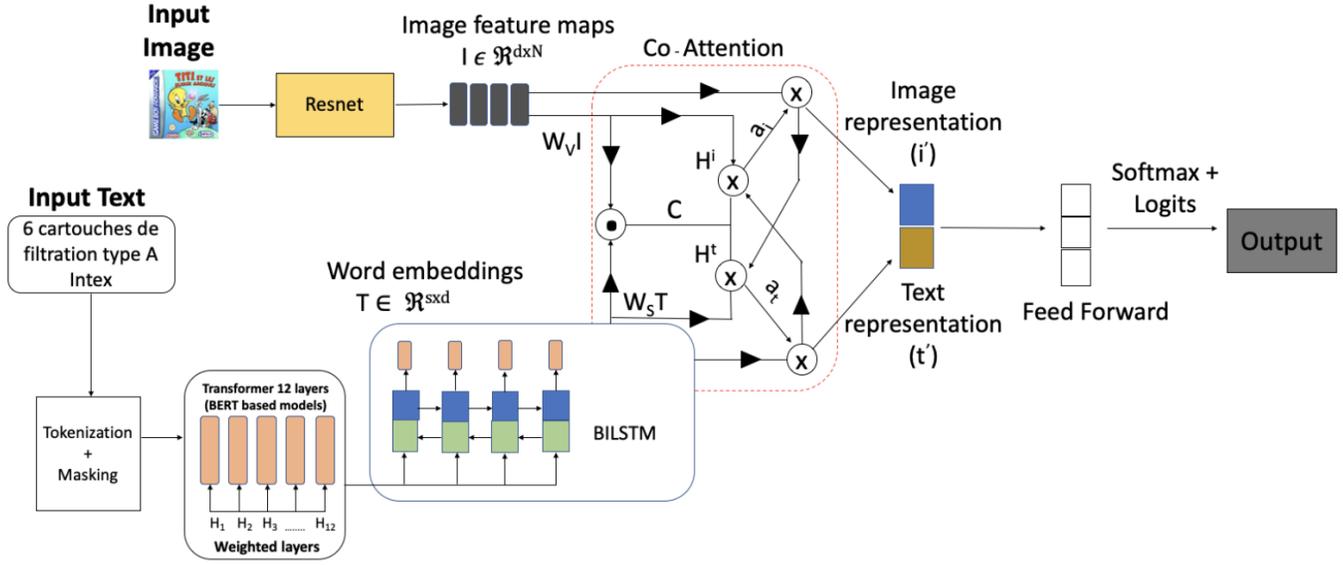

Figure 1: Model architecture representing the Co-Attention technique

To combine the text and images features, we implement the following functions ($Z$):

- Concatenation - Concatenate the image and text embedding features into a single dimension vector.
- Bilinear Transformation - Apply a bilinear transformation to fuse the information from image and text features
- Dot Product - Element wise dot product of image and text features. Text features are passed through a feed forward layer to match the feature dimension of images.

### 3.2 Co-Attention Transformer Model

Figure 1 summarizes our approach. The architecture comprises four important components: 1.Pretrained image model 2.Transformer Language model 3. Bi Directional Long Short Term Memory (BILSTM) 4. Co-Attention. We extract image embeddings using the well-known Resnet model. We encode text using a BERT-based transformer model. The output of the model is a 12-layered hidden state. Common practice to generate embeddings is to extract the top layer of the model. In contrast, we aggregate all the hidden layers using task-specific weightings [16]:

$$O_i = \gamma \sum_{j=0}^{L-1} S_j H_j \quad (2)$$

where

- $H_j$ is a trainable task weight for the $j^{th}$ layer. $\gamma$ is another task trainable task parameter that aids the optimization process
- $S_j$ is the normalized embedding output from the $j^{th}$ hidden layer of the transformer.
- $O_j$ is the output vector.
- $L$ is the number of hidden layers.

The output from the transformer is passed to the BILSTM to learn sequence dynamics. To effectively learn token dependencies, we make use of past and future input features for every specific time step. Back-propagation through time (BPTT) [8] is used while training.

Next, the image and text representations are passed to the co-attention block. We adopt the architecture and method of Lu et al. [11] with one modification. In their work, a hierarchy of attention elements is created between spatial maps and text at 3 levels: words, phrases and sentences. In contrast, our method jointly reasons between words and images. Our rational is that contextual word embeddings from the transformers are a good representation of higher order features [5].

The output from BILSTM is represented as $T \in \mathbb{R}^{s \times d}$, where $d$ is the output dimension and $s$ is the sequence length of words. The output from the pretrained image model after transformation is represented as $I \in \mathbb{R}^{d \times N}$, where d is the feature embedding of images in every image region N.

To compute the co-attention, we follow the steps in Lu et al. [11]. First we calculate the affinity matrix 'C' as given below:

$$C = \tanh(TW_b I) \quad (3)$$

where $W_b$ is the weight parameter.
The affinity matrix is used as a feature to compute the attention maps as shown in the equation below:

$$H^i = tanh(W_V I + (W_S(T)^T)C) \quad (4)$$
$$H^t = tanh(W_S T + (W_V I)C^T) \quad (5)$$

where $W_V, W_S, w_{hi}, w_{ht}$ are weight parameters. The attention probabilities $a^i$ and $a^t$ are used to compute the image and text vectors via a weighted sum over spatial maps (n) and words (s):

$$a^i = softmax(w_{hi}^T H^i) \quad (6)$$
$$a^t = softmax(w_{ht}^T H^t) \quad (7)$$

$$i' = \sum_{n=0}^{N-1} a_n^i I_n \quad (8)$$
$$t' = \sum_{s=0}^{S-1} a_s^i T_s \quad (9)$$

The image and text vectors are concatenated and passed to a neural network with a softmax activation to obtain a vector of class probability scores. Implementation details are provided in Section 4.3

## 4 EXPERIMENTS

### 4.1 Data Description
The data for the task contains 99K products, of which close to 84K items were in the training dataset. Each product in the listing is associated with an image, a French title and an optional description. Figure 2 is a sample of the training data containing multimodal information.

| Integer_id | Title | Description | Image_id | Product_id |
|---|---|---|---|---|
| 0 | Jeep Police - Gevarm-Gevarm | Nan | 1193217616 | 3136702026 |
| 1 | Court Joyeux No√'l En Peluche Taie …… | No√'l en peluche court Taie Sofa | 1323615566 | 4231863665 |
| 2 | Sauna infrarouge Largo ….. | mensions : 150x105x190 cm ou | 1158121321 | 2695198357 |
| 3 | BAGUE POUR LAME SOUS-SOLEUSE G. ET D. | Nan | 1096607258 | 1657064583 |

Figure 2: Example of the multimodal information accompanying the training set

The goal of the task is to predict the product type code (PRC). These codes are numbers associated with a general product name. For example PRC '2280' refers to toys and PRC '1260' refers to electronics. A macro F1-score is used to evaluate the classification accuracy.

### 4.2 Data Preparation & Exploration
To prepare the data for modeling, a first step was text preprocessing, which involved cleaning and processing the textual information. The product description contained html tags, alphanumeric tokens and certain out-of-position product symbols. We stripped these tags as they did not add useful information and may result in improper contextual embeddings from the language models. As a second step, we appended the product description to the title to create a single text corpus that serves as input to our deep learning model.

In figure 3 we plotted the distribution of text length across products. The sequence length describing the product varies significantly with maximum length close to 2000 words. This was relevant to consider, as i) experimenting with various sequence length within the transformer models can help fine tune the model, and ii) Sequence length can be varied to fit different batch sizes within the GPU so as to avoid out-of-memory issues.

We created a train-validation split where we set aside 20% of the train dataset as a validation set. This helped evaluate various models without having to make repeated submissions to the challenge portal.

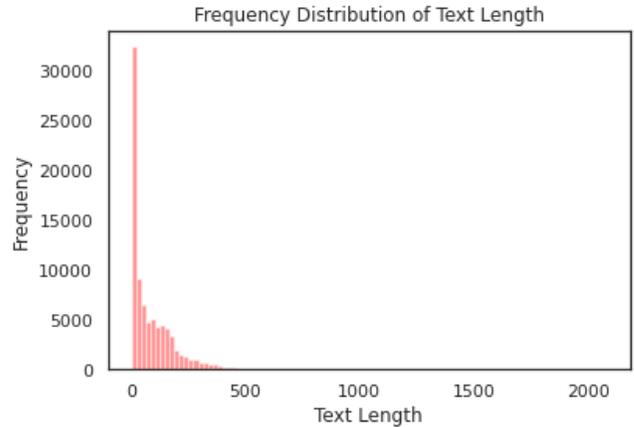

Figure 3: Frequency distribution of the text length

### 4.3 Implementation Details
All our models were implemented in Pytorch and trained on multiple NVIDIA GPUs. For the proposed approach, we used a ResNet-152 [7] or ResNext-101 [20] pre-trained on ImageNet as the backbone network. We discard the final pooling and fully connected layers to produce a feature map of size $16 \times 16 \times 2048$ at the output. $16 \times 16$ is the number of image regions, while 2048 is the dimension of the feature vector of every image region. This can be transformed as into 2048 X 256 matrix. We used CamemBERT, FlauBERT and M-BERT[5] to model textual information. The output of transformer model is passed to a BILSTM with a hidden dimension size of 2048. We initialize the newly added layers using Xavier initialization [6] and optimize with .01 times the learning rate of the base network to avoid over-fitting of the learned features. In all our experiments we used Adam Optimizer [9] with base learning rate 0.0003 and a weight decay of 0.9.

We experimented with varying sequence lengths and batch sizes and set the maximum sequence length to be 510 and the batch size of 16. We sorted the the batches to be of similar sequence length, so as to avoid longer sequences being batched with shorter sequences for faster pre-processing. Following standard protocols, we trained our model over 10 epochs with a categorical cross entropy loss and saved the models with the best F1-score on the validation dataset.

### 4.4 Performance Comparisons
Table 1 shows the performance of our model on the validation set for different combinations of pretrained image and text models. We note that CamemBERT in combination with Resenet152 yields the best performance. The baseline model yielded a macro F1-score of 79.16, demonstrating that the BILSTM and co-attention blocks used in our model are important contributors to model performance.

Table 1: Macro F1 scores of proposed approach for various choices of pretrained models

| Image-Model | CamemBERT | FlauBERT | M-BERT |
|---|---|---|---|
| Resenet152 | 88.78 | 88.52 | 87.06 |
| Resenext101_32x8d | 88.72 | 84.3 | 87.20 |

## 4.5 Ablation Studies

Table 2 shows the performance of our best performing model and the following ablations on the validation dataset:

- **Without weighted layers** - We removed the trainable weighing parameters and considered only the top layer of the transformer as input to the BILSTM.
- **Without BILSTM** - We removed the BILSTM layer, but replaced it with a feed forward layer to match the image embedding dimensions to compute the co-attention.
- **With Image Attention only** - Only image attention was computed and used for classification.
- **With Text Attention only** - Only text attention was computed and used for classification.

Replacing the BILSTM with a linear layer reduced the F1-score significantly which indicates that modeling token dependencies through a transformer alone was not enough.
Without using weighted layers, and only considering the top layer we lose information about the lower layers and notice a drop in the overall performance.

Table 2: Ablation study on our dataset

| Model | F1-Score |
|---|---|
| ResNET152 + CamemBERT | **88.78** |
| ResNET152 + CamemBERT + W/O BILSTM | 76.18 |
| ResNET152 + CamemBERT + W/O weighting layers | 83.53 |
| ResNET152 + CamemBERT + text attention alone | 80.16 |
| ResNET152 + CamemBERT + image attention alone | 86.80 |

## 4.6 Model Ensemble

To further improve performance, an ensemble was created by stacking multimodal models with different base architecture and using an machine learning method that leveraged each individual model's strengths. We gathered probability scores of observations from trained models referenced in table 1, and passed these as an input to a feed forward neural network.
We experimented with different ensemble variations as seen in table 3. The ensemble model that achieved the highest performance was a combination of all six models. Ensembling significantly improved model performance, increasing the F1-score by 2.67 points from the best single model result in table 3.

## 4.7 Error Analysis

We further probed the causes of product label misclassification. We sorted the top-5 most occurring error types and show the distribution in Figure 4. We hypothesize that misclassification is more likely

Table 3: Ensemble Results for various groupings of models using co-attention technique

| Model | F1-Score |
|---|---|
| ReseNET152 + CamemBERT + FlauBERT | 90.40 |
| ResNET152 + Camembert + M-BERT | 90.03 |
| ResNET152 + Camembert + M-BERT + FlauBERT | 90.95 |
| ResNEXT101 + Camembert + FlauBERT | 90.38 |
| ResNEXT101 + Camembert + M-BERT + FlauBERT | 90.88 |
| All 6 models | **91.36** |

to occur amongst products which do not come with a description. To validate this, we conducted statistical t-tests on misclassified samples with and without descriptions. Our hypothesis was defined as:

**Hypothesis $H_0$** (Null hypothesis): *Mean cross-entropy loss is similar for both groups.*

**Hypothesis $H_1$** (Alternate hypothesis): *Mean cross-entropy loss of no - description sample is greater than mean cross-entropy of description sample*

Since the two groups have unequal sample sizes, we used a robust technique known as Welch's t-test:

$$t = \frac{\overline{X_1} - \overline{X_2}}{\sqrt{\frac{S_1^2}{N_1} + \frac{S_2^2}{N_2}}} \quad (10)$$

Table 4 shows that the p-value is statistically significant(<.05), due to which we reject our null hypothesis.

Table 4: t-test statistics

| t-statistic | P-value |
|---|---|
| -6.809 | $1.4 \times 10^{-9}$ |

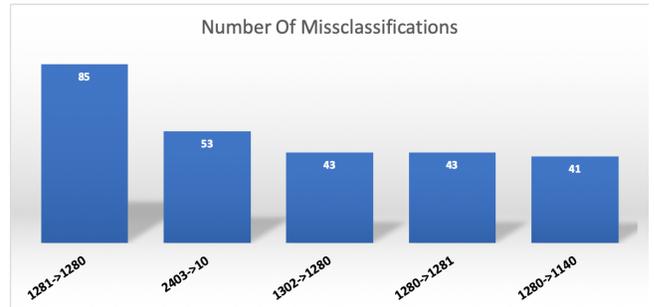

Figure 4: Error distribution of misclassified samples. The X-axis contains the categories that were misclassified. The categories are the predicted class that points to the actual class.

Figure 5 is box plot for the two cases of descriptions being present and absent. We note that error rates are higher in the absence of descriptions.

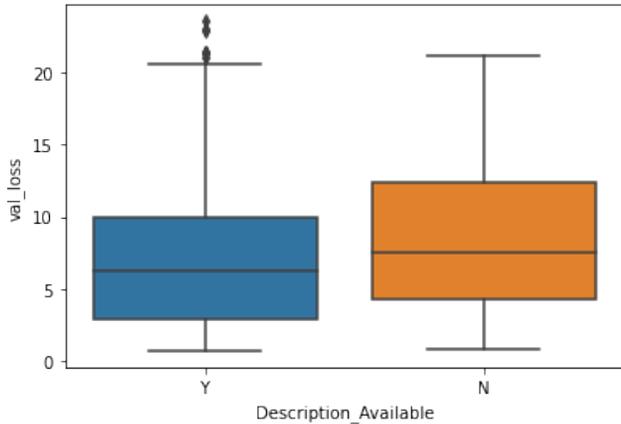

**Figure 5: The boxplot shows the mean validation error is higher when the description is unavailable.**

## 5 CONCLUSION

In this paper we adapted a co-attention transformer model to the task of multimodal product classification. Ensemble learning was used to exploit the strengths of multiple individual models. Different ablation studies and experiments showed that the proposed system performed well on the validation set and was tied for the $2^{nd}$ place on the leaderboard on a hold-out test set. Error analysis revealed that misclassifications were more likely in the absence of product descriptions. One practical challenge we faced is that textual information was available only in French, and thus text-preprocessing tasks and performance comparisons of the proposed technique with existing methods based on English text were nontrivial.

## REFERENCES
[1] Ashutosh Adhikari, Achyudh Ram, Raphael Tang, and Jimmy Lin. 2019. DocBERT: BERT for Document Classification. CoRR abs/1904.08398 (2019). arXiv:1904.08398 http://arxiv.org/abs/1904.08398
[2] Jacob Andreas, Marcus Rohrbach, Trevor Darrell, and Dan Klein. 2015. Deep Compositional Question Answering with Neural Module Networks. CoRR abs/1511.02799 (2015). arXiv:1511.02799 http://arxiv.org/abs/1511.02799
[3] Stanislaw Antol, Aishwarya Agrawal, Jiasen Lu, Margaret Mitchell, Dhruv Batra, C. Lawrence Zitnick, and Devi Parikh. 2015. VQA: Visual Question Answering. CoRR abs/1505.00468 (2015). arXiv:1505.00468 http://arxiv.org/abs/1505.00468
[4] Kan Chen, Jiang Wang, Liang-Chieh Chen, Haoyuan Gao, Wei Xu, and Ram Nevatia. 2015. Abc-cnn: An attention based convolutional neural network for visual question answering. arXiv preprint arXiv:1511.05960 (2015).
[5] Jacob Devlin, Ming-Wei Chang, Kenton Lee, and Kristina Toutanova. 2018. Bert: Pre-training of deep bidirectional transformers for language understanding. arXiv preprint arXiv:1810.04805 (2018).
[6] Xavier Glorot and Yoshua Bengio. 2010. Understanding the difficulty of training deep feedforward neural networks. In Proceedings of the thirteenth international conference on artificial intelligence and statistics. 249–256.
[7] K. He, X. Zhang, S. Ren, and J. Sun. 2016. Deep Residual Learning for Image Recognition. In 2016 IEEE Conference on Computer Vision and Pattern Recognition (CVPR).
[8] Sepp Hochreiter and Jürgen Schmidhuber. 1997. Long short-term memory. Neural computation 9, 8 (1997), 1735–1780.
[9] Diederik P. Kingma and Jimmy Ba. 2015. Adam: A Method for Stochastic Optimization. In International Conference on Learning Representations.
[10] Hang Le, Loïc Vial, Jibril Frej, Vincent Segonne, Maximin Coavoux, Benjamin Lecouteux, Alexandre Allauzen, Benoît Crabbé, Laurent Besacier, and Didier Schwab. 2019. FlauBERT: Unsupervised language model pre-training for french. arXiv preprint arXiv:1912.05372 (2019).
[11] Jiasen Lu, Jianwei Yang, Dhruv Batra, and Devi Parikh. 2016. Hierarchical question-image co-attention for visual question answering. In Advances in neural information processing systems. 289–297.
[12] Louis Martin, Benjamin Muller, Pedro Javier Ortiz Suárez, Yoann Dupont, Laurent Romary, Éric Villemonte de la Clergerie, Djamé Seddah, and Benoît Sagot. 2019. Camembert: a tasty french language model. arXiv preprint arXiv:1911.03894 (2019).
[13] Hyeonseob Nam, Jung-Woo Ha, and Jeonghee Kim. 2017. Dual attention networks for multimodal reasoning and matching. In Proceedings of the IEEE Conference on Computer Vision and Pattern Recognition. 299–307.
[14] Hyeonwoo Noh, Paul Hongsuck Seo, and Bohyung Han. 2015. Image Question Answering using Convolutional Neural Network with Dynamic Parameter Prediction. CoRR abs/1511.05756 (2015). arXiv:1511.05756 http://arxiv.org/abs/1511.05756
[15] Jeffrey Pennington, Richard Socher, and Christopher D Manning. 2014. Glove: Global vectors for word representation. In Proceedings of the 2014 conference on empirical methods in natural language processing (EMNLP). 1532–1543.
[16] Matthew E Peters, Mark Neumann, Mohit Iyyer, Matt Gardner, Christopher Clark, Kenton Lee, and Luke Zettlemoyer. 2018. Deep contextualized word representations. arXiv preprint arXiv:1802.05365 (2018).
[17] Karen Simonyan and Andrew Zisserman. 2014. Very deep convolutional networks for large-scale image recognition. arXiv preprint arXiv:1409.1556 (2014).
[18] Chi Sun, Xipeng Qiu, Yige Xu, and Xuanjing Huang. 2019. How to Fine-Tune BERT for Text Classification? CoRR abs/1905.05583 (2019). arXiv:1905.05583 http://arxiv.org/abs/1905.05583
[19] Damien Teney, Peter Anderson, Xiaodong He, and Anton van den Hengel. 2018. Tips and Tricks for Visual Question Answering: Learnings From the 2017 Challenge. In The IEEE Conference on Computer Vision and Pattern Recognition (CVPR).
[20] Saining Xie, Ross Girshick, Piotr Dollár, Zhuowen Tu, and Kaiming He. 2016. Aggregated Residual Transformations for Deep Neural Networks. arXiv preprint arXiv:1611.05431 (2016).
[21] Huijuan Xu and Kate Saenko. 2015. Ask, Attend and Answer: Exploring Question-Guided Spatial Attention for Visual Question Answering. CoRR abs/1511.05234 (2015). arXiv:1511.05234 http://arxiv.org/abs/1511.05234
[22] Zichao Yang, Xiaodong He, Jianfeng Gao, Li Deng, and Alexander J. Smola. 2015. Stacked Attention Networks for Image Question Answering. CoRR abs/1511.02274 (2015). arXiv:1511.02274 http://arxiv.org/abs/1511.02274